\documentclass[10pt,twocolumn,letterpaper]{article}

\usepackage{iccv}
\usepackage{times}
\usepackage{epsfig}
\usepackage{graphicx}
\usepackage{amsmath}
\usepackage{amssymb}
\usepackage{float}
\usepackage{placeins}
\usepackage{todonotes}

\usepackage[pagebackref=true,breaklinks=true,colorlinks,bookmarks=false]{hyperref}

\iccvfinalcopy 

\def\iccvPaperID{5035} 

\begin{document}

\title{Action Recognition Using Volumetric Motion Representations}

\author{Michael Peven, \hspace{2pt} Gregory D. Hager, \hspace{2pt} Austin Reiter\\
Department of Computer Science, Johns Hopkins University, Baltimore, MD, USA\\
}

\maketitle


\begin{abstract}
Traditional action recognition models are constructed around the paradigm of 2D perspective imagery. Though sophisticated time-series models have pushed the field forward, much of the information is still not exploited by confining the domain to 2D. In this work, we introduce a novel representation of motion as a voxelized 3D vector field and demonstrate how it can be used to improve performance of action recognition networks. This volumetric representation is a natural fit for 3D CNNs, and allows out-of-plane data augmentation techniques during training of these networks. Both the construction of this representation from RGB-D video and inference can be run in real time. We demonstrate superior results using this representation with our network design on the open-source NTU RGB+D dataset where it outperforms state-of-the-art on both of the defined evaluation metrics. Furthermore, we experimentally show how the out-of-plane augmentation techniques create viewpoint invariance and allow the model trained using this representation to generalize to unseen camera angles. Code is available here: \url{https://github.com/mpeven/ntu_rgb}.
\end{abstract}



\section{Introduction} \label{intro}

High-quality, low-cost RGB-D sensors are already commonplace in today's world. Color, depth, and often articulated pose data (3D body configuration) can be collected with ease in real time. The success of action recognition methods depend on how these multi-modal data sources can be used.

Articulated pose data is a popular choice as input to action recognition classifiers as it has the advantage of being an easy-to-use, low-dimensional representation of a human. Pose data lies in 3D Cartesian space, therefore \textit{out-of-plane} data augmentation techniques (3D rotations and translations) can be used to infer novel viewpoints and increase training set size. However, skeletons alone do not provide enough context for activities involving human-object interaction. Furthermore, pose information is often noisy and can only perform as well as the middleware in RGB-D sensors which compute it - by definition limiting performance. Despite the fact that pose is often used for action recognition, it remains to be seen whether this intermediate representation alone is enough to recover a diverse set of human activities from RGB-D data.

The field of action recognition has benefited from the substantial performance improvements of Convolutional Neural Networks (CNNs) over the past decade. We have seen the ability of these networks, trained on millions of images for object recognition, to have considerable generalization when applied to other tasks, including the problem of action recognition in video. However, the performance gains for action recognition have not been commensurate with object detection results - the pre-trained object recognition networks lack the ability to model the temporal structure present in actions. To deal with this, two-stream convolutional networks \cite{simonyan2014two} use pre-trained networks on a single image in conjunction with 2D optical flow over multiple RGB frames to represent motion. Despite these successes, methods using 2D video lack the important 3D spatial representation obtained from RGB-D sensors. 2D data augmentation approaches are limited either to \textit{in-plane} operations or techniques like color jittering that do not represent legitimate physical phenomena. Furthermore, traditional 2D imagery is viewpoint dependent - without context, optical illusions such as “forced perspective” causes an affiliation between overlapping objects that are in fact separated.

In this work, we propose a novel method for classification of actions in video. We design an input representation that takes advantage of 3D motion signals captured from RGB-D sensors without relying on pose data. Using this volumetric input, we investigate an extension to the successful two-stream CNN architecture \cite{simonyan2014two} and experimentally demonstrate how our input representation improves performance over the original 2D formulation. Furthermore, using 3D data augmentation techniques, we show that we can improve performance by generalizing to unseen views. Finally, to capture the long-term structure of human activities, we apply a recurrent neural network which has shown promise in modeling global temporal signals \cite{ng2015beyond}.

To summarize, our contribution is threefold. First, we introduce a novel input to action recognition network using a volumetric representation of 3D motion by projecting 2D optical flow from RGB video into 3D using the z-coordinate captured in a depth-map video. Next, we show that \textit{out-of-plane} data augmentation (translations and rotations) that can be performed over this voxel grid input helps to create view-point invariance. Finally, we obtain state-of-the-art activity classification results using a two-stream CNN network with 3D convolutions over this volumetric representation and further extend it with a long-short term memory network (LSTM) on the output of the convolutional layers. The two-stream network captures spatial and short-term temporal signals from the video while the LTSM models the longer-term temporal structure.

\begin{figure*}
\begin{centering}
\includegraphics[width=0.55\linewidth]{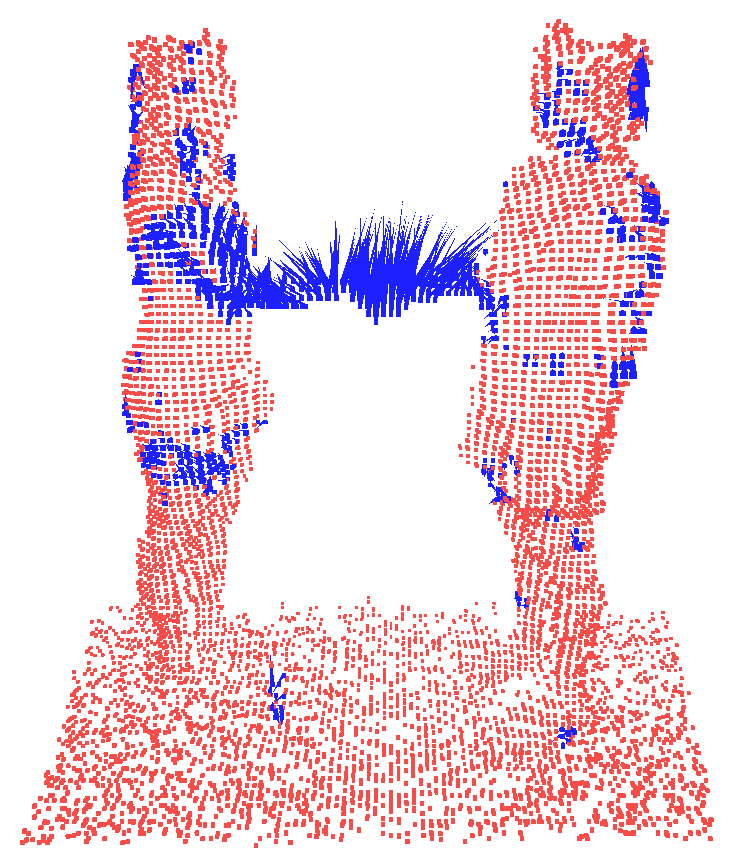}
\caption{Our volumetric representation of motion during the 'shaking hands' action. The red points show the point cloud and the blue vectors represent the 3D motion vectors used as input to the 3D convolutional network. The code to visualize this type of representation from any RGB+D video is provided as an OpenGL application in the released source code.}
\end{centering}
\label{fig:voxel_flow}
\end{figure*}

We have released the code in an open-source repository located at \url{https://github.com/mpeven/ntu_rgb}. It contains everything needed for creating the input representations and training the network on these inputs. Also included is a custom OpenGL application for viewing the input representations. This was built in order to manually validate an interpretable input representation, demonstrated in Figure \ref{fig:voxel_flow}.

The rest of this paper is organized as follows. In Section \ref{relwork}, we discuss related work in the field of action recognition. In Section \ref{methods}, we introduce our method of representing 3D motion from RGB-D data and define our proposed neural network architecture in detail. Implementation and evaluation details are given in Section \ref{results} with a presentation and discussion of our results.

\section{Related Work} \label{relwork}

A broad range of action recognition methods have been studied in the computer vision community over the past few decades. Previous works related to ours fall under the following categories: i) Convolutional networks for action recognition, ii) Long-term temporal modeling for action recognition, and iii) Input representations for action recognition.

\subsection{CNNs for Action Recognition}
The success of CNNs in the image domain has led to a variety of methods when applied toward action recognition. To incorporate the temporal information in video, early work investigated methods involving 3D CNNs \cite{baccouche2011sequential,ji20133d,karpathy2014large,tran2015learning} which perform 3D convolutions over snippets of images to learn spatiotemporal patterns. However, \cite{karpathy2014large} showed similar performance between a 3D CNN over multiple frames and a 2D CNN over a single frame in a large-scale action recognition setting. This result indicates that the spatiotemporal filters in 3D CNNs may not learn a useful temporal signal without a considerable number of training videos. To correct for this, \cite{sun2015human} propose using a 1D temporal convolution to be placed on top of 2D spatial convolutions. Recent work in \cite{carreira2017quo}, show impressive results using 3D CNNs with a deep temporal receptive field. Interestingly, \cite{carreira2017quo} showed that explicitly representing motion using 2D optical flow still improved performance, even when using 3D kernels which are claimed to learn motion signals across video frames.

In this work, we also employ 3D CNNs; however, ours differs in that the three input dimensions are not $XYT$ (where $T$ is the temporal dimension of a video), but $XYZ$, as the input representations we create lie in 3D Cartesian space. Because the convolutions in our 3D CNN are applied over a volume, it is reasonable to assume a symmetric receptive field. Specifically, this means we avoid having to either estimate or experimentally evaluate the receptive field of convolutions in the temporal dimension as is done in \cite{carreira2017quo} and \cite{tran2015learning}, where the choice may not be optimal when generalizing to a new dataset that has different frame-rates or actions that happen at an unfamiliar pace. Following on the success of 3D CNNs for the object recognition domain in \cite{song2016deep}, we investigate its application towards action recognition by using a volumetric motion field.

Our proposed method is most closely related to the two-stream CNNs introduced in \cite{simonyan2014two}. This approach has a spatial CNN, leveraging the success of object recognition networks and pre-trained models and it has a temporal CNN using perspective-based 2D optical flow to represent motion. The substantial success of two-stream approaches has led to numerous successors \cite{carreira2017quo,feichtenhofer2016convolutional,lea2017temporal,liu2017enhanced,wang2015action,wang2015towards} into one of the more effective methods for action recognition to date.

\subsection{Modeling Longer-term Structure}
A limitation of both 3D CNNs and the original two-stream approach is the inability to model the long-term temporal structure present in video. The work introduced in \cite{baccouche2010action} propose recurrent neural networks (RNNs), specifically LSTM cells, to model this long-term structure. This work is expanded upon in \cite{baccouche2011sequential}, where it uses the outputs from a 3D CNN as the input to the RNN. The work in \cite{donahue2015long,lea2017temporal,ng2015beyond,singh2016multi,wang2016temporal} all propose alternative methods of modeling a longer-term signal on top of the outputs of either the original two-stream approach or similar techniques of separating spatial and short-term temporal features. 

\subsection{Input Representations for Action Recognition}
With the emergence of low-cost RGB-D sensors, many datasets for action recognition also include frame-wise depth maps and articulated pose data. A variety of methods look at using more than just the RGB video frames, for example, in \cite{kim2017interpretable,liu2016spatio,liu2017enhanced,rahmani2017learning,shahroudy2016ntu,zhao2017two,zolfaghari2017chained} articulated pose data is used for action recognition; either alone or in addition to RGB and depth video. As the pose output from Microsoft Kinect is calculated using the depth map only, work in \cite{haque2016recurrent,liu2017enhanced} ignore this intermediate output and try to learn a representation with depth-maps only. These works have shown that the 3D representation from pose or depth contains a useful signal; however, state-of-the-art results for action recognition use RGB data in some context (illustrated in Table \ref{table:all_results}).

Similar to our work, \cite{wang2017scene} use scene flow to get a representation of 3D motion for action recognition. However, they constrain this representation to a 2D image, rather than as a volumetric motion field. Furthermore, their work differs from ours as dynamics of these motion fields in the temporal dimension are not explicitly modeled. Although these design choices made in \cite{wang2017scene} are to make use of pre-trained object detectors, the approach we take in this work is to model 3D motion in a voxel grid in order to perform both 3D convolutions and \textit{out-of plane} augmentation techniques directly on this input.

\begin{figure*}[t!]
\begin{center}
\includegraphics[width=\textwidth]{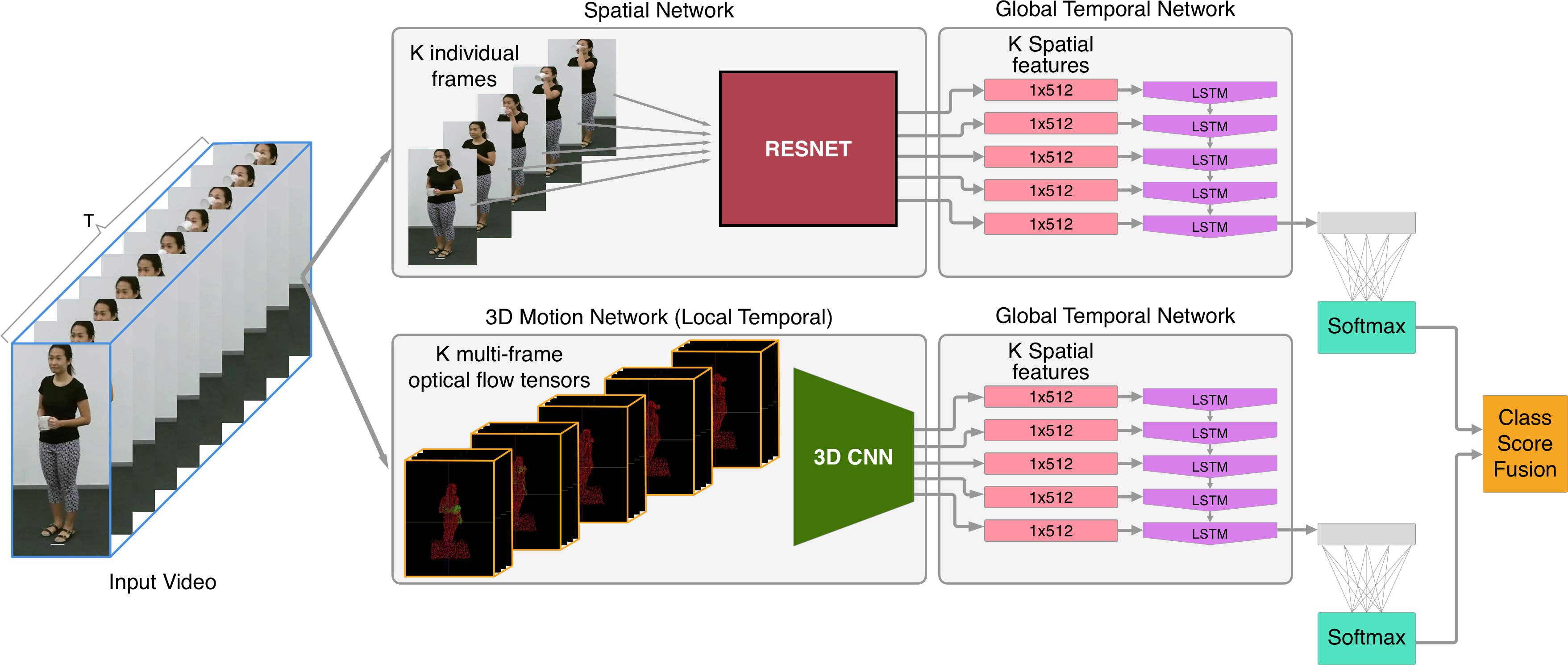}
\vspace*{2mm}
\caption{The network architecture is composed of three main parts: (1) the spatial stream from RGB frames; (2) the local-temporal stream using our 3D motion representation (see Fig. \ref{fig:temporal_network} for more detail on the design of this stream); (3) a global-temporal network to model temporal structure over an entire action.}
\label{fig:full_network}
\end{center}
\end{figure*}

\section{Methods Overview} \label{methods}

We describe here in detail the methods used to transform raw RGB and depth video (from an RGB-D sensor) into a volumetric representation of 3D motion. Formally, the notation we use is as follows: We look at 3 representations of each video $V$, the RGB frames $V_{RGB} \in R ^ {T \times I \times J \times 3}$, the depth frames $V_{depth} \in R ^ {T \times U \times W}$, and the voxelized combination $V_{voxel} \in R ^ {T \times X \times Y \times Z \times 3}$. The height and width of the RGB video and depth video are $I, J$  and $U, W$, respectively. The values $X$, $Y$, and $Z$ are the Cartesian coordinates in the voxel grid and $T$ is the number of frames in the video. The last dimension in the voxel grid is the 3D motion vectors $(dx, dy, dz)$ that we will describe in detail below.

\subsection{3D Voxel Flow} \label{3dmotion}

First, we run dense optical flow over the RGB video to get an estimate of 2D motion over the video. The output of dense optical flow at frame $t$ in $V_{RGB}$ is a displacement vector field, where each vector represents the motion of the scene at pixel $(I_t, J_t)$ to pixel $(I_{t+1}, J_{t+1})$. However, these vectors are in the RGB image system. Using standard camera calibration techniques, we can represent each value in this vector field in terms of the depth image by using the extrinsics between the color and depth cameras and computing a look-up table to map pixels in the RGB image to pixels in the depth image. Finally, given the intrinsic parameters of the depth camera, we can project each pixel in the depth image to its corresponding 3D camera coordinate:

\begin{equation}
\begin{bmatrix}
	X_{camera}\\[1em]
	Y_{camera}\\[1em]
	Z_{camera}
\end{bmatrix}
=
\begin{bmatrix}
\displaystyle\frac{(i - c_x) D_{i,j}}{f_x} \\[1em]
\displaystyle\frac{(j - c_y) D_{i,j}}{f_y} \\[1em]
D_{i,j}
\end{bmatrix}
\end{equation}

\noindent where $D_{i,j}$ is the depth value at pixel coordinates $(i,j)$ and $(c_x, c_y, f_x, f_y)$ are the intrinsic parameters (principle point and focal length, respectively) of the depth camera (note: if the depth camera produces \textit{invalid} data at a particular pixel, we skip it). This allows us to project the 2D vector field from the color image into the 3D camera coordinate system of the depth image. Thus, we have a dense 3D vector field representing motion across frames:
\begin{equation}
\begin{bmatrix}
	dX\\
	dY\\
	dZ
\end{bmatrix}
=
\begin{bmatrix}
X_{t+1} \\
Y_{t+1} \\
Z_{t+1}
\end{bmatrix}
-
\begin{bmatrix}
X_t\\
Y_t\\
Z_t
\end{bmatrix}
\end{equation}

We rescale the 3D motion vectors to fit within a voxel grid. The size of the voxel grid was chosen to be large enough to clearly represent the scene (see Figure \ref{fig:voxel_flow}) but also kept small enough for computational reasons, as discussed below.

\subsection{Voxel Grid Augmentations} \label{augmentations}
The benefit of a volumetric representation is that we can model realistic augmentations, e.g. a change in viewing angle, in comparison to forced 2D perspective augmentation methods. Two of these \textit{out-of-plane} data augmentation techniques to the voxel grid input were investigated. The first method was to translate all of the voxels in the voxel grid in the $X$, $Y$, and $Z$ directions. The second was performing an \textit{out-of-plane} rotation of the motion vectors (rotation about the axis normal to the ground plane).

\begin{figure*}
\begin{center}
\includegraphics[width=0.85\linewidth]{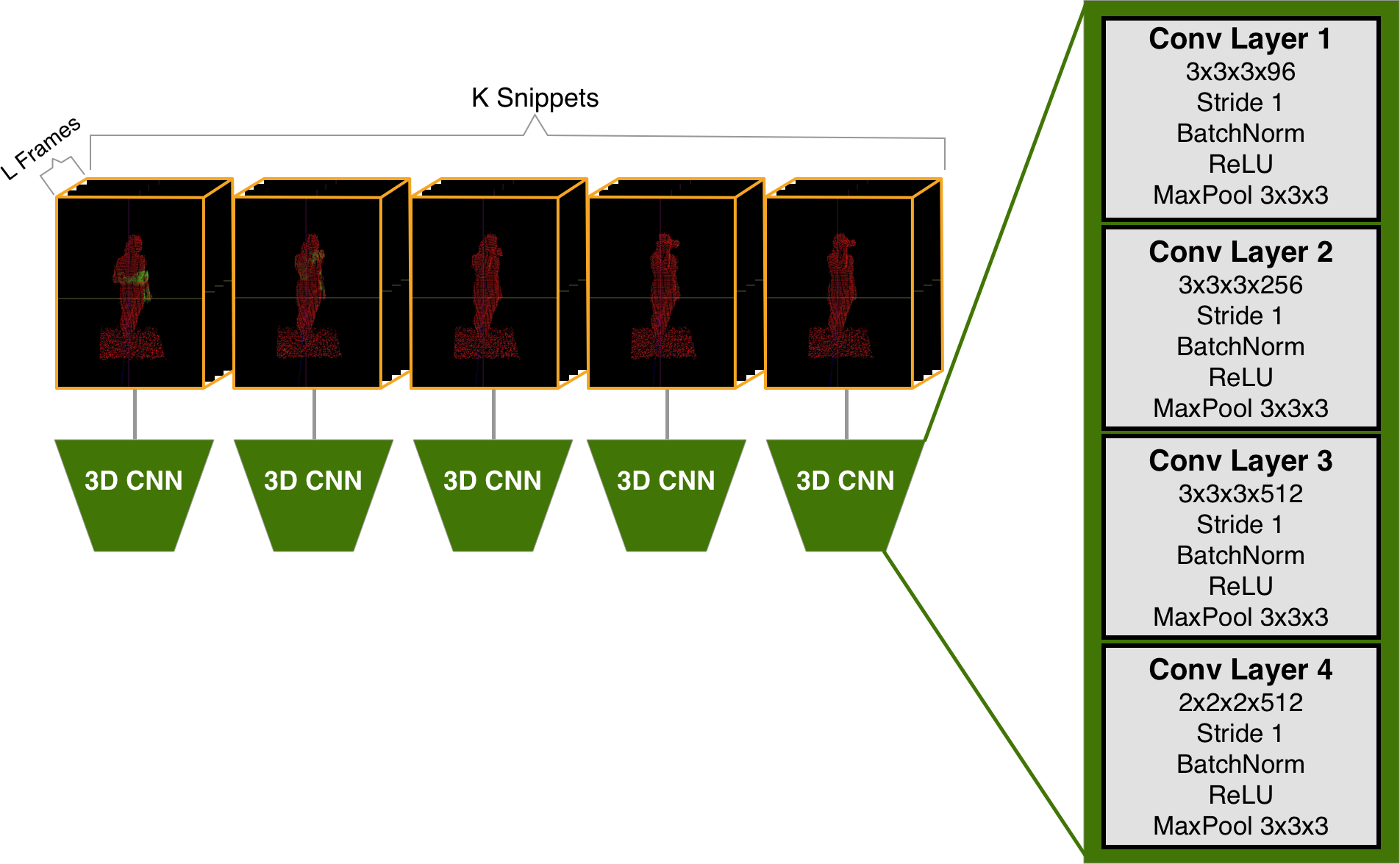}
\vspace*{3mm}
  \caption{The Local-Temporal Stream. This is the network designed for the volumetric motion representation described in Section \ref{3dmotion}. This input is visualized here as images with a red point-cloud and green motion vectors.}
\label{fig:temporal_network}
\end{center}
\end{figure*}

\subsection{3D Two-Stream Model} \label{twostreamintro}
Inspired by the two-stream approach \cite{simonyan2014two} of decomposing a video into a discriminative representation of an action, we wish to segment a video into both spatial and short-term temporal components. This is a powerful approach as it leverages the success of object recognition networks for the spatial stream and explicit modeling of motion in the temporal stream. Furthermore, the spatial stream is important to pick up on the discriminative spatial cues for identical motion patterns, (e.g. mopping vs. sweeping). Equivalently, the temporal motion stream can do the same in the case of identical spatial features (e.g. standing up vs. sitting down). Formally, we would like to represent video $V$ as a snippet $S$, which represents a set of $L$ consecutive frames, i.e. the snippet starting at frame $\tau$ would be represented by $S = (V_{\tau}, V_{\tau+1}, \dots, V_{\tau+(L-1)})$. In this approach, the motion field throughout $S$ represents a localized motion (gesture) and the color image at $S_{L/2}$ represents the median spatial representation of the snippet.

However, this representation of a video is insufficient. It is unable to model the entire structure of an action longer than the number of frames in the video snippet representing the temporal feature \cite{wang2016temporal}. We solve this issue by sampling multiple segments across the entire video. Formally, if $S$ is the input to the temporal network and video $V$ has $T$ frames, we would like to get $K$ sections $S = (S_0, \dots, S_K)$, each of $L$ frames, where the frames for snippet $k$ are chosen as follows:
\begin{align}
	S_k & = (V_{\tau},  V_{\tau + 1}, \cdots, V_{\tau + (L-1)})
\intertext{where:}
	\tau & = k\frac{T-L}{K-1}
\end{align}

\subsubsection{Spatial Stream}

The spatial appearance stream responds to static information in individual video frames. Thus, we build upon the substantial success of residual networks (ResNets) \cite{he2016deep} for object recognition tasks and apply it over the RGB frames sampled from our video. We can leverage the models pre-trained on the millions of images in the ImageNet dataset and fine-tune it towards our task. Training and configuration details are specified in Section \ref{training_details}. As input we use the center image in each sampled section of the video; for snippets $S = (s_0, \dots, s_K)$ of length $L$, input to the spatial stream are the images $(S_{(0,l=L/2)}, S_{(1,l=L/2)}, \dots, S_{(K,l=L/2)})$.

\renewcommand{\arraystretch}{1.25}
\begin{table*}[ht]
\begin{center}
\begin{tabular}{llcc}
\hline\noalign{\smallskip}
Stream & Input & Cross-Subject & Cross-View \\
\noalign{\smallskip}\hline
2D Spatial Stream 			& 2D images			& 81.48\% & 80.60\% \\
\hline
2D Local-Temporal Stream 	& 2D optical flow  	& 71.27\% & 71.09\% \\
3D Local-Temporal Stream 	& 3D motion field	& 78.97\% & 89.58\% \\
\hline
2D 2-Stream 				& 2D images \& 2D optical flow & 88.15\% & 84.24\% \\
3D 2-Stream		 			& 2D images \& 3D motion field & \textbf{89.84\%} & \textbf{94.54\%} \\
\hline
\noalign{\smallskip}
\noalign{\smallskip}
\noalign{\smallskip}
\end{tabular}
\caption{Results of an architecture ablation study on the NTU RGB+D dataset. This table shows the results from the different streams of the network. The first section shows the results from using only the RGB video as input. The second section shows a comparison of the motion stream between a 2D optical flow input and the volumetric input developed here. The final section shows the results of a product of experts voting scheme when combining the output of each stream.}
\label{table:architecture_ablation}
\end{center}
\end{table*}

\subsubsection{Local-Temporal Stream}

The temporal stream of our network is used to learn features from short-term motion patterns. Using the sampling technique described in Section \ref{twostreamintro}, we create $K$ short snippets from an input video. For each snippet, we build the motion representation using the method described in Section \ref{3dmotion} to collect the 3D motion vectors at each frame in the snippet, constructing a voxel grid per snippet, and concatenating the motion vectors in the snippet into their respective voxels. The final representation of a single snippet is $\mathbf{S}_{temporal} \in R ^{3L \times X \times Y \times Z}$, where $X$, $Y$, $Z$ are the dimensions of the voxel grid, and $3L$ represents stacked 3D motion vectors across an entire snippet of length $L$.

We propose a fully 3D CNN (Fig. \ref{fig:temporal_network}) using this representation of motion as input. 3D convolutions are performed in a sliding window style search over each voxel grid. Because these voxel grids are the optical flow based representations of motion, this can be seen as the 3D equivalent of the two-stream temporal network defined in \cite{simonyan2014two} if we choose $K=1$. We consider the effectiveness of a 3D CNN compared to the 2D equivalent, which we describe below.

\subsubsection{Global-Temporal Model}

Modeling the global structure over compositions of the local features computed above is crucial to discriminate between certain actions. For example, taking off a shoe contains the same spatial and temporal patterns as putting on a shoe, albeit in the opposite order. A solution to this, as described in \cite{baccouche2011sequential, donahue2015long, luo2017unsupervised, ng2015beyond} is to model the global structure of an action using recurrent neural networks. We employ an LSTM over the sequences of activations coming from both the spatial and temporal stream. As we do not wish to investigate methods of feature fusion in this work, we use separate LSTMs for the spatial and temporal stream. Therefore, we avoid attempting to model the relationship between space and time, and as can be seen in Figure \ref{fig:full_network}, these networks can ultimately be trained separately. A product of experts voting scheme is used to produce a final classification across streams.

\section{Experiments and Results} \label{results}

The methods introduced are evaluated here. We first present the evaluation dataset. We follow with a description of the implementation and training details. Then, we explore the method used to evaluate the performance of our 3D representation over the 2D baseline and we compare the full two-stream method to current state of the art results. In the end we show in an ablative study the performance gains using 3D data augmentation techniques.

\subsection{NTU RGB+D Action Recognition Dataset}

We evaluate our approach in the NTU RGB+D dataset \cite{shahroudy2016ntu} which to the best of our knowledge, is the largest open-source RGB-D action recognition dataset. This dataset contains RGB, depth, IR, and 3D articulated pose data for each action. There are 56.9 thousand videos split into 60 individual action classes. Of these 60 classes, 40 of which are daily activities, 9 are health related, and 11 are mutual (two-person) actions. The dataset is collected using 3 separate Kinect V2 cameras positioned at angles $-45^{\circ}$, $0^{\circ}$, and $+45^{\circ}$ relative to the subject.
This dataset defines two methods for evaluation, which we follow as described in \cite{shahroudy2016ntu}. The first one is the cross-subject split where a portion of the subjects are assigned to a train split and the rest are reserved for testing. The second one is the cross-view split where videos from two of the camera angles are used for training and the other is used for testing. This corresponds to 40,320 training videos and 16,560 testing videos for the cross-subject split and 37,920 and 18,960 for the cross-view splits.

\renewcommand{\arraystretch}{1.2}
\begin{table*}[!t]
\begin{center}
\begin{tabular}{l|ccc|cc}
\hline
Model & Pose & Depth & RGB & Cross-Subject & Cross-View \\
\hline
Part-Aware LSTM \cite{shahroudy2016ntu}     & X &   &   & 62.9\% & 70.3\% \\
ED-LSTM \cite{luo2017unsupervised}			&   & X &   & 66.2\% & - \\
LSTM with Trust Gates \cite{liu2016spatio} 	& X &   &   & 69.2\% & 77.7\% \\
Res-TCN \cite{kim2017interpretable}         & X &   &   & 74.3\% & 83.1\% \\
DSSCA-SSLM \cite{shahroudy2017deep}			&   & X & X & 74.9\% & - \\
D+S CNN \cite{rahmani2017learning}			& X & X &   & 75.2\% & 83.1\% \\
D-Pose Traversal \cite{weng2018deformable}  & X &   &   & 76.8\% & 84.9\% \\
ESV-CNN \cite{liu2017enhanced} 			    & X &   & X & 80.0\% & 87.2\% \\
Chained-MSN	\cite{zolfaghari2017chained}	& X & X & X & 80.8\% & - \\
Unsupervised \cite{li2018unsupervised}      &   & X & X & 80.9\% & 83.4\% \\
TS-RNNCNN \cite{zhao2017two}			    & X &   & X & 83.7\% & 93.6\% \\
CT-DAN \cite{wang2017cooperative}		    &   & X & X & 86.4\% & 89.0\% \\
Glimpse Clouds \cite{baradel2018glimpse}	& X &   & X & 86.6\% & 93.2\% \\
DA-Net \cite{wang2018dividing}              &   &   & X & 88.1\% & 92.0\% \\
\textbf{Ours}                               &   & X & X & \textbf{89.8\%} & \textbf{94.5\%} \\
\hline
\noalign{\smallskip}
\noalign{\smallskip}
\end{tabular}
\caption{Results of our model compared to state-of-the-art methods on the NTU RGB+D dataset. We present only results from the full 2-stream network, our best performing model.}
\label{table:all_results}
\end{center}
\end{table*}

\subsection{Training and Implementation Details} \label{training_details}

\subsubsection{Sampling parameters} 
As input to the network, $L=5$ frames were chosen to create the volumetric motion representations from $K=10$ segments of the video. In the NTU RGB+D dataset, we found this choice to be reasonable as it captures nearly 60\% of the average video. The size of the voxel grid was chosen to be $54 \times 54 \times 54$. These values were chosen for a number of reasons. Most importantly, when visualized, it is a large enough voxel grid for a human to clearly recognize the actions being performed in the scene. Furthermore, as discussed below, it allows 3D pooling layers after each of the 4 convolution layers to reduce the final output to a vector of size $1 \times N$ where $N$ is the number of convolutions in the final layer.

\subsubsection{CNN parameters} 
We selected the size and configuration of the convolutional layers in our temporal stream to be similar to both \cite{simonyan2014two} and \cite{song2016deep}. These are enumerated on the right hand side of Fig. \ref{fig:temporal_network}. There are 4 layers of 3D convolutions with stride 1, 3D batch-normalization, ReLU activation, and Max Pooling. The kernel sizes were $3^3$, $3^3$, $3^3$, and $2^3$ respectively. The output after the final pooling operation is a vector of size $512$. The global temporal model (seen in Fig. \ref{fig:full_network}) is an LSTM with input size 512 and hidden size 256. Dropout is not applied in this network. The final logits layer has an output equal to the number of classes (60 in NTU RGB+D). The 2D equivalent model, discussed below, has the exact same configuration except with 2D convolutions and 2D batch-normalization. The predictions of the two-stream model uses the product of experts method \cite{hinton2002training} of multiplying each stream's softmax output, however, gradient calculations for backprop were performed before the fusion of the two networks.

\subsubsection{Training details} 
We use cross-entropy loss to optimize our models. The ADAM \cite{kingma2014adam} optimizer is used with a learning rate of 0.001, reducing it by 50\% every 10 epochs. We train all models for 40 epochs or until the training accuracy has saturated. For the spatial stream, the ResNet-18 implementation in Pytorch was used. We use pre-trained ImageNet weights, however the weights of the first half of the network were frozen to avoid overfitting. All experiments were run on a machine with dual Titan-RTX GPUs. The batch size was 8 (videos) when training the motion stream and 128 for the spatial stream. These were chosen because of GPU memory limitations. For training it took on average 33 minutes per epoch for the motion stream and 6 minutes per epoch for the spatial stream (each epoch being a full pass over the training split). Inference over the entire test set took an average of 12 minutes, approximately 23 videos per second. This greatly exceeds real-time performance as it runs at over 1,000 frames per second. In order to determine stopping criteria during training, 5\% of the training samples were reserved as a validation set.

\subsubsection{Data augmentation and pre-processing} 
During training, random translations, as described in Section \ref{augmentations} are applied up to $\pm 10\%$ the total size of the grid. Random \textit{out-of-plane} rotations of voxels and motion vectors are applied at $\pm 30^{\circ}$ about the axis normal to the ground-plane. We examine the performance of these techniques in an ablation study over the NTU RGB+D dataset (Table \ref{table:augmentation_ablation_study}). For the spatial stream over color images, the images need to be pre-processed before being passed through ResNet by resizing the images to (224,224) and normalizing the color values to the same mean and standard deviation. During training, we augment images through random crops, adding random color jitter, and rotating them $\pm 15^{\circ}$. The random parameters of these pre-processing and augmentation techniques are consistent across an individual sample (video), varying only between samples and across training iterations. During inference, no augmentation is applied, only rescaling and normalization. To obtain the 2D optical flow images, the off-the-shelf implementation of \cite{farneback2003two} in the OpenCV toolkit is used.

\subsection{Experimental Results and Analysis}

An ablative experiment designed to evaluate our input representation is presented in Table \ref{table:architecture_ablation}. We first investigate the results from the spatial and temporal stream alone. Additionally, we investigate the 3D representation of motion in our temporal stream with a comparison to the 2D equivalent. We can see that the individual spatial and temporal streams achieve competitive results on their own, with performance rivaling or better than many of the state of the art results presented in Table \ref{table:all_results}. However, the best results come from the full 2-stream model, showing an increase in accuracy of up to 14\% above its respective individual stream. This corroborates that motion and spatial features are discriminative on a disjoint set of features and the full two-stream approach is the most effective.

Using a 3D representation of motion shows a significant performance increase in comparison to the 2D equivalent. The largest increase comes in the cross-view split with an 18.5\% performance increase in classification accuracy. We hypothesize these results show that \textit{out-of-plane} data augmentation techniques with a 3D representation better capture geometric (view-point) invariance than the 2D equivalent. To substantiate these claims, we design an ablative experiment of the augmentation presented in Table \ref{table:augmentation_ablation_study} which we discuss in detail below. Of note, we can see that there is no significant difference between the results of using a 3D representation without data augmentation techniques and its 2D equivalent. This shows that a 3D representation alone is not enough.

\subsection{Comparison with the State of the Art}

Table \ref{table:all_results} presents the results of previous works compared to our proposed 3D two-stream approach. For sake of comparison, we also include the modalities the various models use for training their respective models. Although various modalities were used in the methods we compare against, we felt it fair to compare across all of them, as all outputs from an RGB-D sensor (depth, RGB, and pose) are available for download. These results show that our methodology improves over the previous state-of-the-art results on this dataset.

\subsection{Exploration - Data Augmentation}

We conduct further experiments to understand the difference in performance when using 3D data augmentation techniques. Table \ref{table:augmentation_ablation_study} presents the results of this ablation study. Applying random translations to the voxel grids we see a $5-8\%$ increase in the classification rate of each evaluation metric. Applying random \textit{out-of-plane} rotations to the volumetric representation boosts performance by another $2-9\%$. It is interesting to note that the the relative performance gain on the cross-view split (17.73\%) is much higher than the cross-subject split (7.79\%). Because the cross-view split is from a different camera angle, we believe this illustrates that \textit{out-of-plane} rotations and translations during training help the network become invariant to view-point as evidenced by this significant perfomance gain.

\renewcommand{\arraystretch}{1.2}
\setlength{\tabcolsep}{5pt}
\begin{table}
\begin{center}
\begin{tabular}{lcc}
\hline\noalign{\smallskip}
Augmentation Technique & Cross-Subject & Cross-View \\
\noalign{\smallskip}
\hline
\noalign{\smallskip}
No augmentation 			& 71.18\% & 71.85\% \\
Translations 				& 76.70\% & 80.08\% \\
Translations + Rotations	& \textbf{78.97\%} & \textbf{89.58\%} \\
\noalign{\smallskip}
\hline
\noalign{\smallskip}
\noalign{\smallskip}
\end{tabular}
\caption{Results on NTU RGB+D: augmentation ablation study. This shows classification accuracy when using only the volumetric motion representation (the temporal stream of the network) with the various augmentation techniques applied to the input.}
\label{table:augmentation_ablation_study}
\vspace{-10pt}
\end{center}
\end{table}
\setlength{\tabcolsep}{1.4pt}

The volumetric representation created in this work is from a single RGB-D camera. This is limiting because the type of video, referred to as 2.5D, can only capture partial surfaces from the depth-map, i.e. the camera-facing surface. Without complete surface reconstruction, the data augmentations through virtually rotating and translating the scene is limiting due to missing information. However, our results show a clear benefit towards applying this technique nonetheless, evidenced by the significant performance gain shown in Table \ref{table:augmentation_ablation_study}.

\section{Conclusions}

We propose a method for activity recognition using a volumetric representation of 3D motion. The method achieves state of the art performance on both the cross-subject and cross-view evaluation metrics of the NTU RGB+D dataset. A novel representation of motion is created from RGB-D video by projecting a dense optical flow field, calculated over RGB frames, into a 3D voxel grid using the corresponding depth map. A two-stream convolutional network is applied over short snippets of video and an LSTM is used to model the temporal structure over the snippets. In the two-steam network, the spatial stream uses a pre-trained object recognition network for the RGB frames and for the temporal stream we define a 3D CNN formulation over the volumetric 3D motion field. In our experiments, we show that this 3D representation outperforms the equivalent 2D representation. Furthermore, we show that the \textit{out-of-plane} data augmentation techniques that are possible over a 3D representation can significantly improve performance. Future work will investigate approaches to fuse the temporal and spatial stream to allow learning of complementary filters across streams.

{\small
\bibliographystyle{ieee}
\bibliography{citations.bib}
}

\end{document}